\documentclass[letterpaper, 10 pt, conference]{ieeeconf}  

\IEEEoverridecommandlockouts                              
\usepackage[free-standing-units=true]{siunitx}
\usepackage{mathtools}
\usepackage{booktabs,tabularx,caption}
\usepackage[usestackEOL]{stackengine}
\usepackage{array}
\usepackage{makecell}

\renewcommand{\baselinestretch}{0.978}

\usepackage{amsmath,amssymb,stmaryrd,mathtools}
\usepackage[noend]{algpseudocode}
\usepackage{acronym}
\usepackage{verbatim}
\usepackage{booktabs}
\usepackage{siunitx}
\usepackage{graphics}
\usepackage{graphicx,caption}
\usepackage[keeplastbox]{flushend}
\usepackage{suffix}
\usepackage{xstring}
\usepackage{xparse}
\usepackage{expl3}
\usepackage{mathrsfs}
\usepackage{tabularx}
\usepackage{makecell}
\usepackage{array}
\usepackage{hyperref}
\usepackage[capitalize]{cleveref}
\usepackage[ruled,linesnumbered]{algorithm2e}

\usepackage{subcaption}
\usepackage{wrapfig}

\usepackage{tikz}
\usetikzlibrary{arrows,backgrounds,calc}
\usepackage{relsize}
\usepackage{float}
\usepackage{kantlipsum} 
\usepackage{lipsum}
\usepackage{stfloats}

\usepackage{todonotes}
\usepackage{soul}
\definecolor{smoothgreen}{rgb}{0.7,1,0.7}
\sethlcolor{smoothgreen}

\RequirePackage{luatex85}
\usepackage{pgfplots}
\pgfplotsset{compat=newest}
\pgfplotsset{every axis legend/.append style={%
		cells={anchor=west}}
}
\usepgfplotslibrary{polar}
\usetikzlibrary{arrows}
\tikzset{>=stealth'}

\definecolor{C1}{rgb}{0.0, 0.447, 0.741}
\definecolor{C1_light}{rgb}{0.0, 0.6032388663967612, 1.0}
\definecolor{C2}{rgb}{0.85, 0.325, 0.098}
\definecolor{C3}{rgb}{0.929, 0.694, 0.125}
\definecolor{C4}{rgb}{0.494, 0.184, 0.556}
\definecolor{C5}{rgb}{0.466, 0.674, 0.188}
\definecolor{C6}{rgb}{0.301, 0.745, 0.933}
\definecolor{C7}{rgb}{0.635, 0.078, 0.184}

\usepgfplotslibrary{groupplots}

\usetikzlibrary{shapes.geometric, arrows}

\tikzstyle{startstop} = [rectangle, rounded corners, minimum width=2cm, minimum height=1cm,text centered, draw=black, fill=none]
\tikzstyle{arrow} = [thick,->,>=stealth]
\usepackage[style=ieee, url=false, doi=false, natbib=true, mincitenames=1, maxcitenames=1]{biblatex}

\newcommand{\ph}[1]{{\textbf{#1}:}}

\title{Occlusion-Aware Crowd Navigation Using People as Sensors}

\author{Ye-Ji Mun$^1$, Masha Itkina$^2$, Shuijing Liu$^1$, and Katherine Driggs-Campbell$^1$
\thanks{This project was supported in part by the Ford-Stanford Alliance and a gift from Mercedes-Benz Research \& Development North America, and in part by the National Science Foundation under Grant No. 2143435.}
\thanks{$^{1}$Ye-Ji Mun, Shuijing Liu, and Katherine Driggs-Campbell are with the Electrical and Computer Engineering
Department, University of Illinois at Urbana-Champaign, USA. Email: {\tt \{yejimun2, sliu105, krdc\}@illinois.edu}.}%
\thanks{$^{2}$Masha Itkina is with the Aeronautics and Astronautics Department, Stanford University, USA. Email: {\tt \{mitkina\}@stanford.edu}.}%
}

\begin{document}

\maketitle
\thispagestyle{empty}
\pagestyle{empty}


\begin{abstract}
Autonomous navigation in crowded spaces poses a challenge for mobile robots due to the highly dynamic, partially observable environment. 
Occlusions are highly prevalent in such settings due to a limited sensor field of view and obstructing human agents. Previous work has shown that observed interactive behaviors of human agents can be used to estimate potential obstacles despite occlusions. We propose integrating such social inference techniques into the planning pipeline. We use a variational autoencoder with a specially designed loss function to learn representations that are meaningful for occlusion inference. This work adopts a deep reinforcement learning approach to incorporate the learned representation into occlusion-aware planning. In simulation, our occlusion-aware policy achieves comparable collision avoidance performance to fully observable navigation by estimating agents in occluded spaces.
We demonstrate successful policy transfer from simulation to the real-world Turtlebot 2i. To the best of our knowledge, this work is the first to use social occlusion inference for crowd navigation. Our implementation is available at \url{https://github.com/yejimun/PaS_CrowdNav}.
\end{abstract}

\section{Introduction}
\label{sec:intro}

Navigating in a pedestrian-rich environment is an important yet challenging problem for a mobile robot due to deficiencies in perception. In cluttered settings, spatial occlusions are inevitable due to obstructing human agents and a limited sensor field of view (FOV). Existing crowd navigation methods often neglect occlusions and assume complete knowledge of the environment is provided~\cite{van2011reciprocal, chen2017decentralized}. When deployed in the real-world, these algorithms only consider the detected or observed human agents for collision avoidance. As a result, collisions may occur when occluded human agents suddenly emerge on the robot's path. However, under similar limitations, humans can safely navigate as they instinctively reason about potential risks. Humans are able to complement their limited sensing capabilities using insights from their past experiences as well as their understanding of social norms (e.g. keeping an appropriate distance from others)~\cite{helbing1995social}. Similar to humans, planning policies should be able to intelligently make inferences in occluded regions to safely navigate partially observable environments. 

Previous literature in autonomous driving has proposed successful occlusion-aware planning algorithms~\cite{bouton2019safe, isele2018navigating}, but the setting considered is an inherently structured environment such as an intersection. In crowd navigation, the mobility of human agents is unrestricted resulting in highly diverse behaviors~\cite{rehder2018pedestrian}, and, thus, making occlusion reasoning more challenging. Prior works~\cite{hara2020predicting, mitkina2021, afolabi2018people, sun2019behavior} demonstrate that missing environmental information can be inferred by observing other people's interactive behaviors. For example, a human slowing down or stopping abruptly may imply the presence of an obstacle in its path as humans tend to follow the principle of least action and keep their speed constant~\cite{helbing1995social}. 
\begin{figure}[t]
	\centering
	\includegraphics[width=0.47\textwidth]{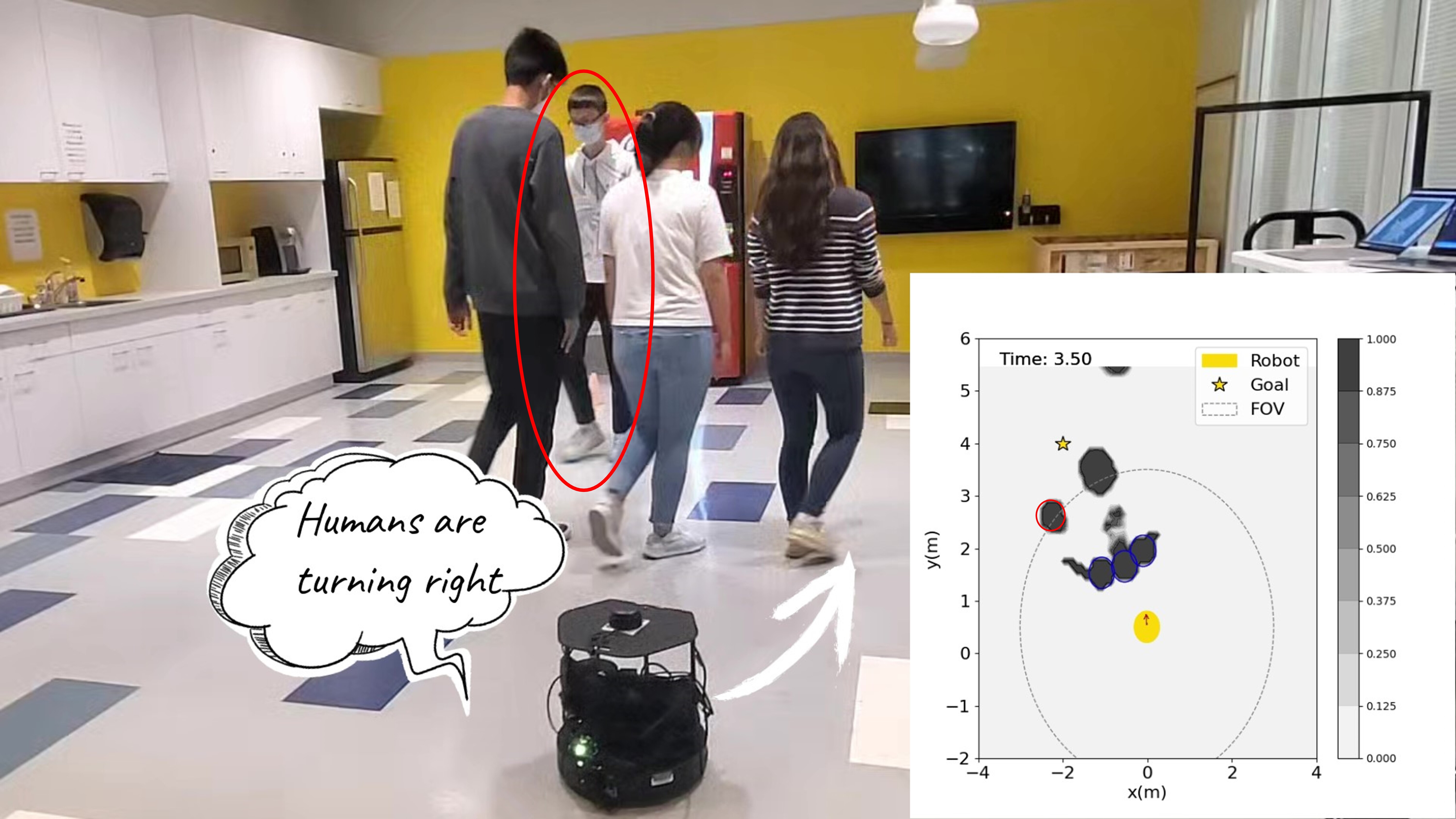}
	\caption{\small Turtlebot2i reasoning about a potential occluded human (red circle) based on interactive behaviors of the observed humans. Our PaS inferred OGM in occluded regions is shown on the right. 
 } 
 \vspace{-0.4cm}
    \label{fig:intro_turtlebot}
\end{figure}

Our work is inspired by a growing body of literature on social inference.
\citet{afolabi2018people} first coin the term `People as Sensors' (PaS) and demonstrate how occluded pedestrians can be inferred by observing human drivers' reactions. 
This work employs occupancy grid maps (OGMs)~\cite{elfes1989using} for representing agents and the environment as they do not require prior environment knowledge and can handle an arbitrary number of agents in the scene~\cite{itkina2019dynamic}.
\citet{mitkina2021} scale the PaS framework to multi-agent social inference in driving scenes by posing the occlusion inference task as a sensor fusion problem.
In this work, we explore the use of PaS in unstructured crowd navigation settings to estimate the location of occluded, freely traversing human agents.
We also go beyond inference by integrating the social inference features into planning and analyzing how our enhanced perception pipeline can improve collision avoidance strategies.

We propose incorporating this social inference mechanism into a deep reinforcement learning (RL) algorithm for robust navigation in a partially observable, crowded environment. We train our policy network end-to-end with an occlusion inference module to augment the incomplete perception. For occlusion inference, we employ a variational autoencoder (VAE)~\cite{kingma2014vae} architecture to encode interactions between human agents into a low-dimensional latent space using specialized loss terms. The RL policy network takes the latent representation as input, which enables the robot to proactively avoid occluded agents. Simulation results show significant improvement in partially observable navigation with our occlusion inference technique. We demonstrate successful policy transfer to the real-world Turtlebot2i.

\ph{Contributions} 
(1) We propose a deep RL framework for map-based crowd navigation that can make occlusion-aware action plans for a partially observable, cluttered environment. 
(2) We integrate a VAE into the deep RL algorithm that is trained using specialized loss terms to extract features for occlusion inference. 
(3) We demonstrate that the joint learning of the occlusion inference and path planning modules results in targeted map estimation that can handle temporary and long-term occlusions enabling proactive collision avoidance.

\section{Related Works}
\label{sec:related}

\ph{Occlusion Inference}
Occlusion inference strategies must be adapted to the occlusion type (i.e. partial vs. full and temporary vs. persistent) and the nature of the environment. Several studies use semantic segmentation to inpaint the unobserved portions of partially occluded objects~\cite{lu2020semantic, purkait2019seeing}. During temporary occlusions, previously observed objects can be hallucinated from memory using recurrent neural networks (RNNs) and skip-connections~\cite{ebert2017self, dequaire2018deep}. \citet{wang2020roll} hallucinate static objects using a long short-term memory (LSTM)~\cite{hochreiter1997long} network and an auxiliary matching loss. Inspired by this approach, we also incorporate a matching loss, but our algorithm performs high-level reasoning for dynamic humans in the presence of long-term occlusions. A new line of work proposed reasoning about persistently fully occluded dynamic agents using the reactive behaviors of observed human agents~\cite{hara2020predicting, afolabi2018people, mitkina2021, amirian2021we}. \textcite{amirian2021we} extract statistical patterns from past observations to estimate the probability of human occupancy in occluded regions of crowded scenes. \citet{afolabi2018people} infer the presence of an occluded pedestrian in a crosswalk from the reactive behaviors of an observed driver. \citet{mitkina2021} generalize this idea to multiple drivers as `sensors' by employing sensor fusion techniques. We also use the social behaviors of human agents to inform occlusion inference of temporarily and persistently fully occluded agents. We incorporate the interactive features in an RL framework to improve navigation. 

\ph{Planning Under Occlusions}
A partially observable Markov decision process (POMDP)~\cite{pomdp1995} is often used to explicitly consider hidden states when planning under occlusions~\cite{bouton2019safe, bouton2017belief}. However, these approaches require the number of occluded agents to be pre-specified, and are intractable with a large number of agents. Deep RL methods have the capacity to capture complex features without requiring prior knowledge of the environment. \citet{liang2021crowd} demonstrate sim-to-real steering in densely crowded scenes using deep RL. To handle occlusions, the robot learns to make sharp turns to avoid suddenly emerging pedestrians from occluded regions. We present a means to anticipate such occluded agents using observed social behaviors in crowds, resulting in smoother robot trajectories. 
\citet{wang2020learning} construct a deep RL algorithm to achieve 3D map-based robot navigation in static, occluded environments. Following this line of work, we propose a map-based deep RL approach that handles occlusions, while navigating highly dynamic environments. 

\ph{Crowd Navigation}
Classical crowd navigation techniques like social force models~\cite{helbing1995social} and velocity-based methods~\cite{van2011reciprocal, van2008reciprocal, snape2011hybrid}
follow predefined reaction rules to avoid collisions (e.g. taking the right side of the path to avoid other agents). However, these reaction-based approaches can be short sighted and over-simplify pedestrian strategies for collision avoidance~\cite{chen2017socially, chen2020relational}. Other works perform long horizon obstacle avoidance by first predicting human agent trajectories and then finding a feasible path that safely avoids the human agents~\cite{aoude2013probabilistically, kretzschmar2016socially, trautman2013robot}. These trajectory-based methods are known to suffer from the robot freezing problem in dense crowds where a feasible path may not be found. Learning-based approaches have been shown to more closely imitate human-like behaviors by learning implicit features that encode social behaviors~\cite{chen2017socially}. 
Pair-wise interactions between agents are often learned to reason about a dynamic environment and perform collision avoidance~\cite{chen2017decentralized,liu2020decentralized}.
In such methods, the complexity grows with the number of agents in the scene. Additionally, only visible, fully detected agents are typically considered. In our algorithm, we employ OGMs to compactly represent an arbitrary number of agents and learn the mutual influence between agents simultaneously.
\section{Problem Statement}
\label{sec:problem statement}

We consider a crowd navigation task where a mobile robot encounters occlusions caused by some agents obstructing other agents from view or by a limited FOV. The robot's goal is to safely avoid all nearby human agents despite limited visibility and efficiently navigate to its target location. 

We formulate the partially observable interactions between agents as a model-free RL problem with continuous state and action spaces, $S$ and $A$. At each time $t$, the robot in state $s_t\in S$ takes an action $a_t \in A$ given an observation $o_t \in \mathcal{O}$. The policy $\pi:\mathcal{O} \rightarrow A$ directly maps the observed state $o_t$ to an action $a_t$ that maximizes the future discounted return:

\vspace{-0.14cm}
\begin{equation} \label{eq:value}
\begin{aligned}
V^\pi (s_t) = \sum_{k=t}^{\infty}\gamma^k R (s_{k}, a_{k}, s'_{k}),
\end{aligned}
\vspace{-0.14cm}
\end{equation}
where $R(s,a, s')$ is the reward function and $\gamma$ is the discount factor. We assume that the human agents' movements are not influenced by the robot. This assumption is common for crowd navigation as it prevents the robot from achieving collision avoidance effortlessly (i.e. the human agents circumvent the robot while the robot marches straight toward its goal)~\cite{liu2020decentralized}. Since our aim in this work is to investigate if the robot can employ occlusion inference to prevent collisions in occluded settings, this assumption encourages the robot to actively reason about the presence of occluded agents.

We employ OGMs to represent the environment map surrounding the robot from a bird's-eye view as shown in \cref{fig:architecture}. As collisions are unlikely to occur with distant agents, we consider a local OGM around the robot for policy learning. We generate two local OGMs centered around the robot at time $t$: a ground-truth OGM $G_t \in \{0,1\}^{H \times W}$ and an observation OGM $O_t \in \{0,0.5,1\}^{H \times W}$, where $H$ and $W$ are the OGM height and width, respectively. The ground-truth OGM $G_t$ captures the true occupancy information for all visible and occluded obstacles, as indicated with free~($0$) or occupied~($1$) values in each cell. The observation OGM~$O_t$ includes the region of uncertainty ($0.5$) referred to as the occluded space, in which some human agents may be hidden. 

Within the observation map $O_t$, we constrain the robot's FOV to a smaller circular region around the robot, as shown in \cref{fig:architecture}(a).
This FOV simulates the limited range of a ray-based sensor (e.g. LiDAR). Outside of this circular area is unknown space. 
We assume that the whole observation map $O_t$ is the critical area for collision avoidance. Thus, the objective of our algorithm is to infer the unknown occupancy in $O_t$ caused by dynamic obstructions and a limited FOV. 
We build the observation map using ray tracing to find detected obstacles. Then, the cells behind the detected obstacles from the robot's perspective are treated as occluded. We focus on estimating fully occluded obstacles, and, as such, assume partially occluded agents can be detected in the OGM. 

\section{Methodology}
\label{sec:methods}
\begin{figure*}[t]
	\centering
	\includegraphics[width=\textwidth, height=0.29\textheight]{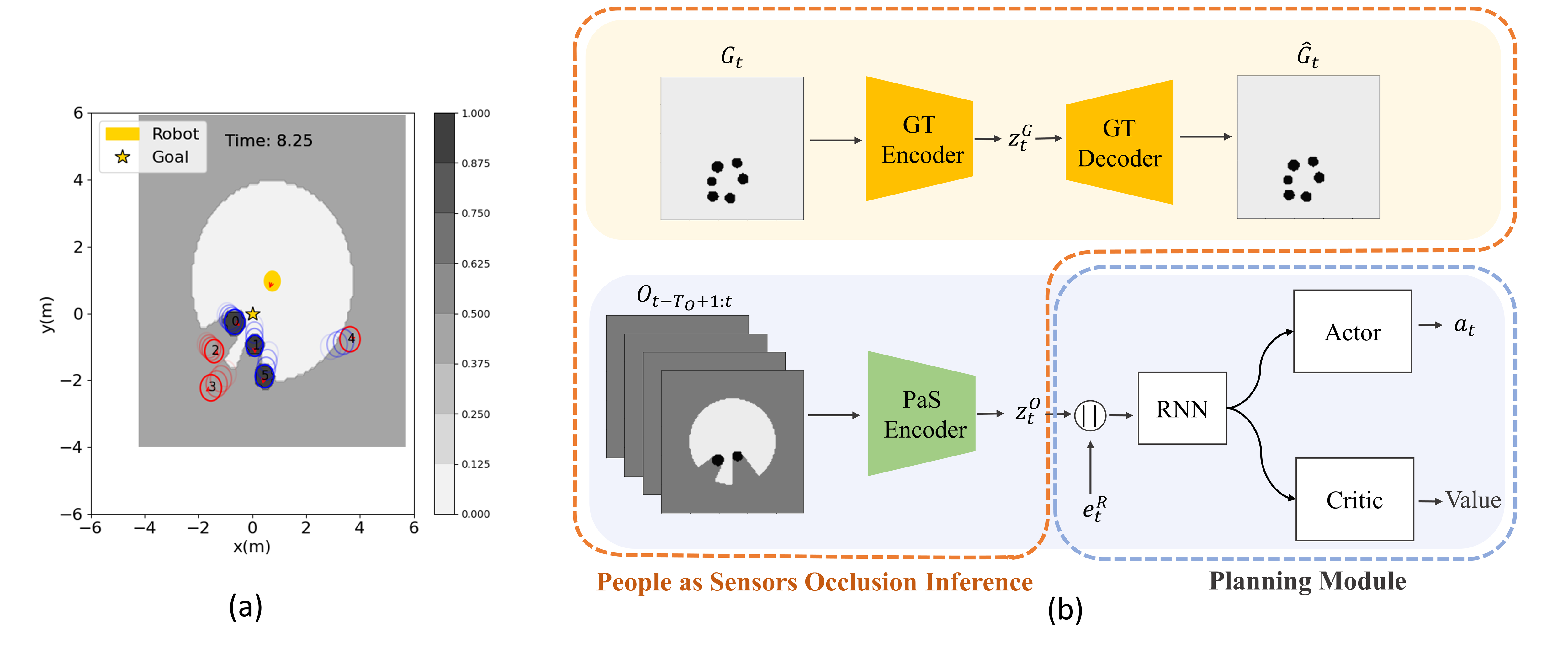}
  \vspace{-0.7cm}
	\caption{\small (a) Illustration of our crowd navigation environment with three observed humans (blue circles) and three occluded humans (red circles). The robot is shown as a yellow circle with its goal as a yellow star. The robot's observation is represented as the OGM. (b)~Our occlusion-aware crowd navigation algorithm consists of a PaS occlusion inference module and a path planning module. The GT-VAE~(top) is first pre-trained with ground-truth OGM data. Under the supervision of the GT-VAE, the PaS encoder (bottom) and the actor-critic network in the planning module are jointly optimized for occlusion inference and proactive collision avoidance.} 
    \label{fig:architecture}
      \vspace{-0.4cm}
\end{figure*}

Our approach comprises an occlusion inference module for map estimation and a path planning module for navigation as shown in~\cref{fig:architecture} (b). We train the two components in an end-to-end manner so that our perception system and decision-making protocol jointly inform and improve each other.

\subsection{Occlusion Inference Module}
Our occlusion inference module handles both short-term and long-term occlusions. We define short-term occlusion as the case when a human agent who is not observable at the current time step has been seen previously within a certain observation time window $T_O$. An agent who has not been seen throughout this period constitutes a long-term occlusion. We aim for the robot to learn to infer these occluded agents as it learns its navigation policy. The joint optimization between the map estimation and planning modules facilitates a perception branch tuned for navigation and a planner optimized for environmental uncertainties.

Short-term occlusion is handled by using a sequence of past observation OGMs $O_{t-T_0+1:t}$ as input to our network. Based on these observations, an agent's trajectory pattern can be extrapolated to its current location in the grid. Estimating the presence of long-term occlusions requires higher-level reasoning. For such unseen agents, we use observed agents' interactive behaviors to infer the presence of occluded agents, following our past work~\cite{mitkina2021}. We extract these interactive features from $O_{t-T_0+1:t}$ using two VAEs: a ground-truth VAE (GT-VAE) and a People as Sensors VAE (PaS-VAE). The GT-VAE is pre-trained and provides supervision for the PaS-VAE that is trained with the policy network.

\ph{Ground-Truth VAE}
The GT-VAE acts as a supervisor for the occlusion inference task in our method. It encodes the most recent omniscient OGM $G_t$ 
into a compact $1$D latent vector $z_t^G$. We train GT-VAE using the ELBO loss~\cite{kingma2014vae}, consisting of an $\ell_2$-norm reconstruction loss and a Kullback–Leibler divergence (KL loss).
As the omniscient OGM~$G_t$ is not available during test time, the GT-VAE is only used during training. To accelerate policy training, the GT-VAE is pre-trained independently from the RL policy and its parameters are frozen during policy learning.

\ph{People as Sensors VAE}
The PaS-VAE takes in a sequence of observations ($O_{t-T_O+1:t}$) that only has partial information of the environment map as shown in \cref{fig:architecture}~(b). The sequence of observations allows us to learn from observed social behaviors to inform occlusion inference. For example, a human agent slowing down or taking a detour to avoid another nearby agent can be gleaned from the sequence. To enable the PaS-VAE to learn the missing information from these spatiotemporal interactions, we take advantage of the pre-trained GT-VAE, which contains complete spatial map data. Assuming it is sufficiently well-trained, the GT-VAE can serve as a supervisor model and provide full map information through its latent encoding $z_t^G$. Thus, we perform occlusion inference by matching the encodings between the GT-VAE and the PaS-VAE using our PaS matching loss:
\vspace{-0.1cm}
\begin{equation} \label{eq:PaSloss}
\begin{aligned}
    L_{\text{PaS}} =\Vert z_t^G-z_t^O\Vert^2_2, 
\end{aligned}
\end{equation}
where $z_t^O$ in the PaS-VAE encodes the estimated map, including occluded agent information, given the OGM inputs $O_{t-T_O+1:t}$. Intuitively,
the PaS-VAE can augment its environmental knowledge by associating the social behaviors in its limited observation with the current, full spatial features encoded in the supervisor encoder output $z_t^G$. Unlike GT-VAE, the PaS-VAE directly influences policy learning as its latent vector 
$z_t^O$ is passed to the policy network, and the two networks are jointly optimized.

We also consider standard VAE loss terms such as the KL loss to regularize the PaS-VAE and an adapted reconstruction loss:
$L_{\text{est}} = \Vert G_t - \hat{O}_t \Vert^2_2$ to minimize the $\ell_2$-distance between the ground-truth OGM $G_t$ and our estimated OGM $\hat{O}_t$. Instead of optimizing a separate PaS decoder, we use the frozen GT-VAE decoder to output the estimated OGM~$\hat{O}_t$ for the estimation loss $L_{\text{est}}$ as the encodings between GT-VAE and PaS-VAE should be similar by the PaS loss. However, we found the estimation loss $L_{\text{est}}$ degrades the navigation performance in favor of more accurate map reconstruction, thus we did not use it for the results in \cref{sec:results}.

\subsection{Policy Learning}
To enable occlusion-aware path planning, the planning module takes as input the latent encoding $z_t^O$ provided by the PaS-VAE. Along with the encoded map, the network takes the robot's state vector as input to help localize itself during navigation: $s^R = (x_R, y_R, v_x, v_y)$ where $(x_R, y_R)$ is the robot's relative position to the goal and $(v_x, v_y)$ is its velocity. The latent encoding $z_t^O$ and the robot's state embedding $e^R_{t}$ are concatenated and fed into the planner as shown in \cref{fig:architecture}~(b). 

For the planning module, we adopt the PPO algorithm~\cite{schulman2017proximal}. The module comprises an RNN and two multi-layer perceptrons (MLPs), corresponding to the actor and critic networks. The critic network approximates the value function, while the actor network updates the policy distributions suggested by the critic. Our PPO implementation is adapted from~\cite{liu2020decentralized, kostrikov2018pytorch}. However, the planning module can be replaced with any model-free 
RL algorithm. 

\section{Experiments}
\label{sec:exp}
\begin{figure*}[ht]
	\centering
	\includegraphics[width=0.93\textwidth]{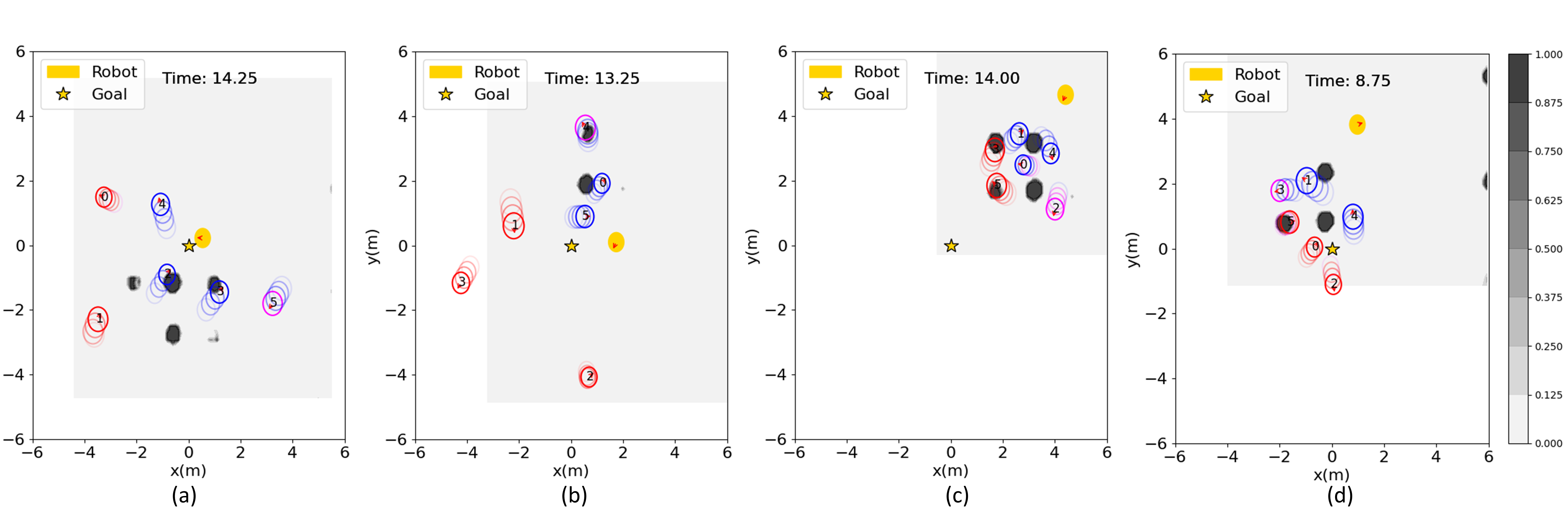}        
	\caption{\small Illustration of our occlusion inference performance. 
 The figure shows human agent trajectories in our sequential observation input for \SI{1}{\second} and the reconstructed OGMs $\hat{O}_t$ from our PaS encoding. The observed and occluded humans are denoted as blue and red circles, respectively. If an agent is temporarily occluded but has been seen in the past \SI{1}{\second}, it is denoted in magenta. In the OGM $\hat{O}_t$, higher occupied probability cells are darker in shade. The estimated OGMs from our approach show properties that promote better navigation. In \cref{fig:PaSestimation}(a), the PaS encoding favors estimation of occluded agents that may pose a potential danger to the robot such as an approaching agent (humans 2 and 3) 
 rather than human 4 who is moving away at a distance from the robot. In \cref{fig:PaSestimation}(b), despite fewer observed interactions at the boundary of the robot's FOV, temporarily occluded human 4 is successfully inferred by extrapolating from previous observations. In \cref{fig:PaSestimation}(c), the slow speed of observed human 0 can be used to infer occluded agents ahead like humans 3 and 5. In \cref{fig:PaSestimation}(d), human 3 making an avoidance maneuver (i.e. a turn) provides insight that occluded obstacles like human 5 may be to its left. Our algorithm successfully provides insight for better crowd navigation in the presence of occlusion based on observed social behaviors.}
	\label{fig:PaSestimation}
  \vspace{-0.4cm}
\end{figure*}

\subsection{Simulation Setup}
\cref{fig:architecture}~(a) shows our simulation environment. The scene includes six human agents, each modeled as a circle with a radius in $[0.3,0.4)$ \si{\meter}. Each agent's initial position is set along a circle with a \SI{4}{\meter} radius and offset with random noise sampled from a uniform distribution over $[-1,1)^2$. The goal positions are symmetric about the origin. All human agents are controlled by Optimal Reciprocal Collision Avoidance (ORCA)~\cite{van2011reciprocal} with a maximum speed in $[0.5,1.5)$ \si{\meter/\second}. The human agents reciprocally influence each other's trajectories but do not react to the robot as discussed in \cref{sec:problem statement}.

The robot is modeled as a circle with a \SI{0.3}{\meter} radius. We set the robot's goal at the origin, where the interactions between agents are the richest and occlusions are likely. The robot's starting position is randomly sampled from a \SI{10}{\meter}$\times$\SI{10}{\meter} square centered at the origin, with a minimum of \SI{6}{\meter} from the goal. 
The robot is assumed to be a holonomic system with a maximum speed and acceleration of \SI{2}{\meter/\second} and \SI{1}{\meter/\second\squared}, respectively. Centered at the robot, the observation OGM $O_t$ is a \SI{10}{\meter}$\times$\SI{10}{\meter} square, and the robot's FOV is a circle with a \SI{3}{\meter} radius. The OGM resolution is \SI{0.1}{\meter}.

\subsection{Implementation Details}
\ph{Reward Function}
Similar to~\cite{liu2020decentralized}, we reward the robot for reaching its goal and penalize (near) collisions with other agents. To encourage the robot to move toward the goal, it receives a positive (negative) reward for getting closer to (further away from) its goal. 
The reward function is:
\begin{equation}
  R(s_t, a_t) =
    \begin{cases}
       10 & \text{if }d_{t}^{g}< r_{robot}\\
      -5 & \text{if }d_{t}^{min} < 0\\
      2.5(d_{t}^{min}-0.25) & \text{if }0 \leq d_{t}^{min} < 0.25\\
      2(-d_{t}^{g}+d_{t-1}^{g}) & \text{otherwise}
    \end{cases}       
\end{equation}
\noindent where $s_t=(O_t,s^R_t)$ is the observed state, $d_{t}^{g}$ is the robot's distance from its goal, $r_{robot}$ is the robot's radius, and $d_{t}^{min}$ is the $l_2$ distance of the robot to its closest agent at time $t$.

\ph{Network Structure}
The GT-VAE and PaS-VAE consist of $5$ convolutional blocks, containing $4\times4$ convolutional layers, Batch Normalization~\cite{ioffe2015batch}, and a rectified linear unit (ReLU), as well as two linear layers to obtain the mean and variance for the latent variable. While the GT-VAE takes in a single OGM $G_t$ at the current time step, the PaS-VAE takes a sequence of $T_O=4$ observation OGMs $O_{t-T_O+1:t}$ as input, containing \SI{1}{\second} of observation history. Our policy network consists of a single layer for the robot state embedding, an RNN with a hidden state of size $128$, and two subsequent $4$-layered MLPs for the actor and critic networks.

\ph{Baselines}
We compare the navigation performance of our method to baseline models trained with different perception levels: omniscient ground-truth (GT) and limited sensor observation (OBS) without occlusion estimation. 
We compare our model-free approach with ORCA~\cite{van2011reciprocal}, a state-of-the-art (SOTA) model-based method. We adapted ORCA to limited FOV for fair comparison. To ablate our results, we evaluate model-free baselines that have a similar network structure as our PaS-VAE in \cref{fig:architecture} but exclude some training components. The GT/OBS-FE baselines take an OGM ($G_t$ or $O_t$) at the current time step as input and are trained without 
the PaS matching and KL losses. For these baselines, our PaS encoder performs feature extraction (FE) for policy learning 
without map estimation. The Seq-GT/OBS-FE baselines take sequential OGMs ($O_{t-T_0+1:t}$ or $G_{t-T_0+1:t}$) as input to explore the effect of incorporating observed behaviors into path planning. The Seq-GT-VAE serves as the oracle model. It is the same as our PaS-VAE but instead takes a sequence of ground-truth OGMs $G_{t-T_0+1:t}$ as input.

\ph{Training}
We trained all models for $15\times 10^6$ time steps. The training details are adapted from~\cite{liu2020decentralized, kostrikov2018pytorch, schulman2017proximal}.  
Each policy gradient update used 180 frames from six episodes. We collected $4$ frames per second. The networks are optimized using Adam~\cite{kingma2014adam} with 
a $10^{-5}$ learning rate. The gradient norm is clipped at $0.5$. The KL coefficient was $0.00025$. We used $100$ episodes for validation and $500$ episodes for testing.

\section{Results}
\label{sec:results}
In this section, we evaluate our occlusion inference module and compare our end-to-end occlusion-aware planning policy to the baselines. 
We demonstrate the benefits of reasoning about occlusions in crowds in terms of safety and efficiency. 

\ph{PaS Occlusion Inference} 
To evaluate the ability of our algorithm to infer occupancy in occluded regions based on observed agent behaviors, we visualize the PaS latent encodings as estimated OGMs using the GT decoder in \cref{fig:PaSestimation}.
We observe that the estimated obstacles do not perfectly align with the agents due to end-to-end training with the deep RL planner. Instead, the estimated OGM captures important features for navigation, rather than for perfect reconstruction.

\cref{fig:PaSestimation}(a) shows that the estimated occupancy from the PaS encoding emphasizes inference of occluded agents who may interfere with the robot's path (humans $2$ and $3$) rather than agents who are moving away at a distance from the robot (human $4$). 
Since map estimation is trained end-to-end with the policy network, the PaS-VAE learns to draw attention to the risks that directly affect the robot's policy objectives. 

Our method successfully estimates temporarily occluded 
pedestrians, such as human 4 at the edge of the robot's FOV in 
\cref{fig:PaSestimation}(b).
Often agents that are farther away interact less with observable agents, limiting the social cues.
However, the sequence of observation inputs to the PaS-VAE and the RNN in the policy module provide sufficient temporal information for the perception pipeline to reasonably estimate human~$4$. 

Lastly, our algorithm infers plausible locations for fully occluded agents who have not previously been detected or seen for a long time. \cref{fig:PaSestimation}(c) and (d) show examples of observed interactive behaviors, such as slowing down or performing an evasive maneuver, that enable PaS-VAE to infer the presence of occluded agents in the vicinity. 
Thus, jointly optimizing the map estimation and planning modules results in a perception branch that is tuned for navigation and able to handle temporary and long-term occlusions.

To quantitatively evaluate map estimation, we use image similarity (IS)~\cite{image_similarity}. IS assesses the structural similarity of OGMs using Manhattan distance (lower is better). IS has been shown to be more representative of OGM quality than a cell-wise metric (e.g., mean squared error)~\cite{toyungyernsub2021double,lange2021attention}. We evaluate IS \{occupied, free, occluded, total (std. error)\} on the estimated OGM $\hat{O}_{t}$ relative to the ground-truth OGM $G_t$. We obtained IS values of $\{43.5, 0.0440, 398, 441 (2.81)\}$ for our PaS estimated map and of $\{27.7, 14.8, 400, 443 (6.78)\}$ for the observation OGM. While our PaS map can be imprecise in occupancy estimation according to the occupied IS score, the effective free space estimation (low free IS score) facilitates our method's efficient path planning. Rather than mapping all observable agents precisely, our task-aware approach prioritizes the observable and inferred occluded agents that directly affect the robot's path to the goal.

\begin{figure*}[t!]
	\centering
	\includegraphics[width=0.75\textwidth]{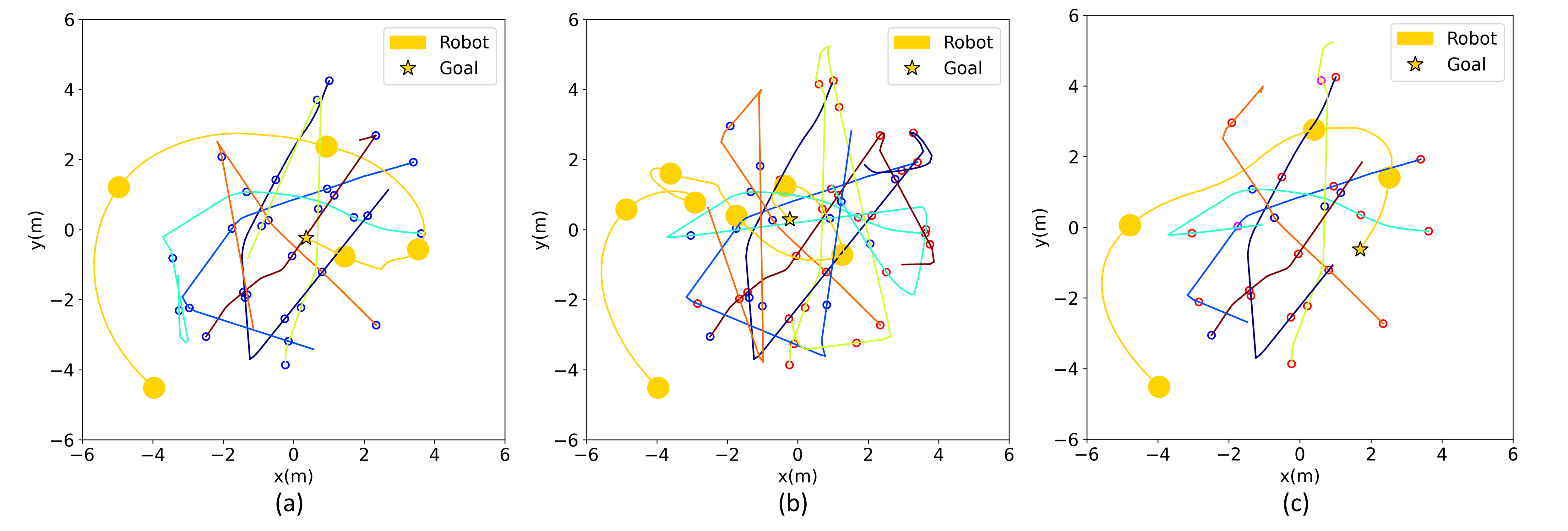}
         \vspace{-0.2cm}
	\caption{\small Trajectories of the robot (yellow) and human agents for (a) Seq-GT-VAE, (b) OBS-FE, and (c) our PaS-VAE. Our approach takes a comparable route to the oracle Seq-GT-VAE. While OBS-FE is highly reactive to unexpected agents resulting in sharp turns, our algorithm reaches the goal using a more efficient and smooth trajectory.}
    \label{fig:episode_traj}
    \vspace{-0.4cm}
\end{figure*}

\ph{Occlusion-Aware Collision Avoidance}
\cref{table:crowdnav_result} shows our quantitative navigation results. The SOTA model-based baseline, ORCA, degrades in performance significantly when the robot's FOV is limited. This performance drop is inevitable for model-based approaches when the model assumptions break (i.e. reciprocal collision avoidance in ORCA assumes full visibility). We employ our model-free, learning-based approach to intelligently reason about a partially observable environment based on observed social context.

\begin{table}[t!]
\centering
\caption{\small Navigation results in terms of success rate, collision rate, discomfort rate, navigation time, and path length. The success rate is the collision-free goal reaching rate. Timeouts, when neither success nor collision occur, are not included. The best (second best) performance, excluding the oracle Seq-GT-VAE, is denoted in bold (underlined). Our PaS algorithm outperforms the limited view OBS baselines and performs comparably to the omniscient GT baselines.} 
\scalebox{0.95}{
\begin{tabularx}{0.5\textwidth}{@{}p{1.9cm} p{0.8cm}p{0.8cm}p{1.2cm}p{1.0cm}p{0.8cm}@{}} 
\toprule
Method &  \Centerstack{Success\\rate} & \Centerstack{Collision\\rate} & \Centerstack{Discomf.\\rate} & \Centerstack{Nav.\\time (s)} & \Centerstack{Path\\length (m)} \\
\midrule
\midrule
GT-ORCA & \hfil 0.7 & \hfil 0.3 & \hfil \underline{0.02}& \hfil 17.54 & \hfil \hfil \underline{8.06} \\
OBS-ORCA & \hfil 0.44 & \hfil 0.56 & \hfil 0.04 & \hfil \textbf{13.47} & \hfil \textbf{6.59} \\
\midrule
OBS-FE& \hfil 0.79 & \hfil 0.21 & \hfil \underline{0.02} & \hfil 16.15 & \hfil 13.8 \\
GT-FE & \hfil 0.83 & \hfil 0.17 & \hfil \textbf{0.01} & \hfil 17.31 & \hfil 13.33 \\ 
Seq-OBS-FE &\hfil  0.82 &\hfil 0.18 &\hfil 0.02 &\hfil 16.39 & \hfil13.68  \\
Seq-GT-FE &\hfil 0.87 &\hfil 0.13 & \hfil \textbf{0.01} & \hfil 17.56 & \hfil 16.53\\ 
PaS-VAE (Ours)  &\hfil \textbf{0.91} & \hfil \textbf{0.09} & \hfil \underline{0.02}& \hfil \underline{13.91} & \hfil12.26\\ 
\midrule
Seq-GT-VAE & \hfil 0.97 & \hfil 0.03 & \hfil 0.00 & \hfil 14.85 & \hfil 13.39 \\
\bottomrule
\end{tabularx} 
}
\vspace{-0.5cm}
\label{table:crowdnav_result}
\end{table}

We ablate the performance contribution of each component in our algorithm. Our PaS-VAE success rate is comparable to the oracle baseline (Seq-GT-VAE) and performs much better than the limited sensor view baselines (OBS-FE and Seq-OBS-FE) by a maximum of a $12\%$ margin. Comparing the sequential (Seq-GT/OBS-FE) and single (GT/OBS-FE) OGM input baselines, we see that past observations help improve planning. Despite the RNN in all baseline policy networks, the sequential observation input is still helpful to learn the human agent dynamics and improve collision avoidance. Aside from better behavior modeling, PaS occlusion inference components in our approach (i.e. the PaS matching and KL losses) allow the joint system to use past observations coupled with intelligent inference to improve navigation in partially observable settings. Interestingly, despite the Seq-GT-FE and the oracle baseline taking the same omniscient sequential input, the oracle surpasses the other by a $10\%$ success rate with our PaS occlusion inference components.

We observe that excessive map information induces slower navigation times and more conservative robot behavior.
The omniscient view baselines (GT-ORCA, GT-FE, and Seq-GT-FE) took longer to reach the goal than the limited view baselines (OBS-ORCA, OBS-FE, and Seq-OBS-FE). Our PaS-VAE has the fastest navigation time and traverses the shortest path to the goal among the learning-based approaches. We attribute the improved efficiency to our map estimation module being trained end-to-end to achieve good performance in crowd navigation. We note that the PaS policy can be more aggressive than the omniscient baselines as indicated by the discomfort rate. In a dense crowd, keeping a large comfort distance with other agents is not always desirable as it often results in the robot freezing problem~\cite{trautman2015robot}. In the real world, human agents are generally cooperative and willing to allow passage to others~\cite{trautman2015robot}. Thus, our PaS policy results in more human-like crowd navigation, while remaining highly safe in terms of collisions. 

\cref{fig:episode_traj} shows qualitative navigation results. For both Seq-GT-VAE and our method, the robot approaches the goal in a smooth trajectory. In contrast, due to the limited FOV, OBS-FE is more reactive, sharply changing direction to avoid collision with unforeseen agents, and results in a longer navigation path and time. By estimating agents in occluded regions, our method anticipates potential upcoming pedestrians and plans accordingly for smoother and more efficient navigation that is akin to the oracle policy. 

\section{Real-World Experiment}
We implemented our occlusion-aware policy on a Turtlebot 2i as shown in \cref{fig:intro_turtlebot}. We used a 2D RPLIDAR A3 for localization and person detection. We used the Gmapping~\cite{gmapping} ROS package for the environment map and the adaptive Monte Carlo localization (AMCL)~\cite{pfaff2006robust} package to get the robot's state. To detect pedestrians, we adopted the DR-SPAAM algorithm~\cite{jia2021domain}. We trained our PaS model as described in \cref{sec:exp} but with four simulated humans and using unicycle dynamics considering the Turtlebot 2i's speed limits. In real-world crowd navigation, when the robot observed three humans suddenly turn right, it successfully estimated an occluded human agent on the left and safely avoided collision. 
Our video demonstration is available \href{https://yejimun.github.io/PaS_CrowdNav_video.html}{here}.
\section{Conclusion}
\label{sec:conclusion}
We proposed an occlusion-aware deep RL planner that proactively avoids collision with agents in crowded, partially observable settings. We use social behaviors of human agents as the key insight to make occlusion inferences. We extract these interactive features from a sequence of observations using a VAE and our specialized matching loss. This augmented perception information is trained in tandem with policy learning in an end-to-end manner.
Our policy achieves performance similar to a policy with ground-truth map information and significantly outperforms baselines with no occlusion inference. We also demonstrate successful policy transfer to a Turtlebot2i platform in a real-world setting. As future work, we aim to extend our occlusion inference module to occlusion-aware map prediction and test it on a real pedestrian data set with a larger number of humans. 

\section*{Acknowledgement}
We would like to thank Haonan Chen, Kaiwen Hong, and Tianchen Ji for help implementing the localization and person detection algorithms for the real-world experiment. We sincerely appreciate members in the Human-Centered
Autonomy Lab who participated as human agents in the real-world experiments.




\clearpage
\printbibliography

\end{document}